\documentclass[8pt]{article}
\usepackage[margin=20mm]{geometry}

\usepackage{amsmath,amssymb,amsfonts}
\usepackage[ruled]{algorithm2e}
\usepackage{graphicx}
\usepackage{textcomp}
\usepackage{xcolor}
\usepackage{url}
\usepackage{hyperref}
\usepackage{booktabs}
\usepackage{tabularx}
\usepackage{bm}
\usepackage{color}
\usepackage{xcolor}
\usepackage{csquotes}

\usepackage[backend=biber,
            style=numeric,
            sorting=none,
            maxbibnames=5,
            giveninits=true,
            doi=true,
            url=true,
            isbn=false]{biblatex}
            
\def\BibTeX{{\rm B\kern-.05em{\sc i\kern-.025em b}\kern-.08em
 T\kern-.1667em\lower.7ex\hbox{E}\kern-.125emX}}

\newcommand{\mathvect}[1]{\boldsymbol{#1}}

\begin{document}

\title{Real-Time Black-Box Optimization for Dynamic Discrete Environments Using Embedded Ising Machines}

\author{
Tomoya Kashimata, Yohei Hamakawa, Masaya Yamasaki, Kosuke Tatsumura\\
\small Corporate Laboratory, Toshiba Corporation, Kawasaki 212-8582, Japan\\
\small Corresponding author: Tomoya Kashimata (e-mail: tomoya1.kashimata@toshiba.co.jp)
}
\date{}

\maketitle

\begin{abstract}
Many real-time systems require the optimization of discrete variables.
Black-box optimization (BBO) algorithms and multi-armed bandit (MAB) algorithms perform optimization by repeatedly taking actions and observing the corresponding instant rewards without any prior knowledge. Recently, a BBO method using an Ising machine has been proposed to find the best action that is represented by a combination of discrete values and maximizes the instant reward in static environments. In contrast, dynamic environments, where real-time systems operate, necessitate MAB algorithms that maximize the average reward over multiple trials. However, due to the enormous number of actions resulting from the combinatorial nature of discrete optimization, conventional MAB algorithms cannot effectively optimize dynamic, discrete environments. Here, we show a heuristic MAB method for dynamic, discrete environments by extending the BBO method, in which an Ising machine effectively explores the actions while considering interactions between variables and changes in dynamic environments. We demonstrate the dynamic adaptability of the proposed method in a wireless communication system with moving users.
\end{abstract}

\section{Introduction}\label{sec_intro} 
Real-time systems repeatedly take actions to control environments and receive some rewards as the results of the actions.
There are two kinds of rewards: instant reward and average reward. The former is the reward corresponding to an individual action, while the latter is the average of instant rewards over all repetitions. Consider problems to maximize either type of rewards through sampling cycles. The sampling cycle is a combination of taking an action and observing an instant reward for the action, where the reward depends on the action taken in the cycle and does not depend on the previous actions.
Black-Box Optimization (BBO) algorithms correspond to instant reward maximization problems, while Multi-Armed Bandit (MAB) algorithms correspond to the average reward maximization problems.
In MAB algorithms, all actions taken during the entire sampling process are evaluated, while only the best action is evaluated in BBO algorithms.
In both algorithms, balancing exploration and exploitation is important for achieving higher rewards.
Exploration is the process of searching through untried actions to find better ones, while exploitation is the process of utilizing the best action already taken or a similar action to obtain a high reward.

Actions can be represented by combinations of multiple discrete variables, where the number of possible actions increases exponentially with the number of variables.
For example, in the machine operation scheduling problem (Job-shop scheduling problem), an action contains binary variables representing which operation should be performed at a specific time on a machine \cite{schworm2023}.
Combinatorial MAB algorithms do not consider the interaction between variables, i.e., they assume that the reward is given by a linear function of the variables~\cite{talebi16, combes15, cesa12}.
Complex environments where the reward depends on the combination of discrete values make MAB/BBO algorithms computationally challenging.
It is important to select the next action from the enormous number of actions efficiently by utilizing the combinatorial structure.
Kitai \textit{et al.} have proposed ``Factorization Machine with Quantum Annealing (FMQA),'' a BBO algorithm for discrete environments that considers interactions using an Ising machine \cite{kitai20}.
Ising machines can quickly find approximate solutions that minimize (or maximize) quadratic discrete functions which are difficult to solve exactly in practical time.
In the FMQA, a quadratic surrogate model that approximates the environment is constructed by a machine learning method.
Then, the Ising machine finds the next action that maximizes the surrogate model, i.e., the best action expected to yield the highest instant reward.
The FMQA, thus, can effectively find the next action from the enormous number of possible actions.
The FMQA and its derivatives have been applied to material discovery \cite{kitai20, inoue22, doi23, couzinie24}, hyper-parameter search for simulation \cite{xiao24}, matrix compression \cite{kadowaki22}, and engineering problems \cite{matsumori22, hida24}.

Unlike the assumption of stationarity in BBO algorithms, the environment in practical real-time systems may change over time.
Depending on the rate of change, MAB algorithms are classified as stationary MAB algorithms (not changing), dynamic MAB algorithms (gradually changing) \cite{dacosta08, gupta11, nobari19}, and adversarial MAB algorithms (changing arbitrarily at each sampling cycle) \cite{Auer95}.
Dynamic MAB algorithms have practical use cases and are more optimizable than adversarial MAB algorithms.
Studies on dynamic MAB algorithms for complex environments have already been conducted \cite{xu24, heliou23}; however, they cannot guarantee rewards that are sufficiently close to those of the best actions when the environment constantly changes.
In addition, algorithms that require time proportional to the number of possible actions to determine the next action become impractical when the number of possible actions increases exponentially.
MAB algorithms that efficiently explore actions using an Ising machine have not yet been proposed.

\begin{figure}[tb]
\centering
\includegraphics[width=17.2 cm]{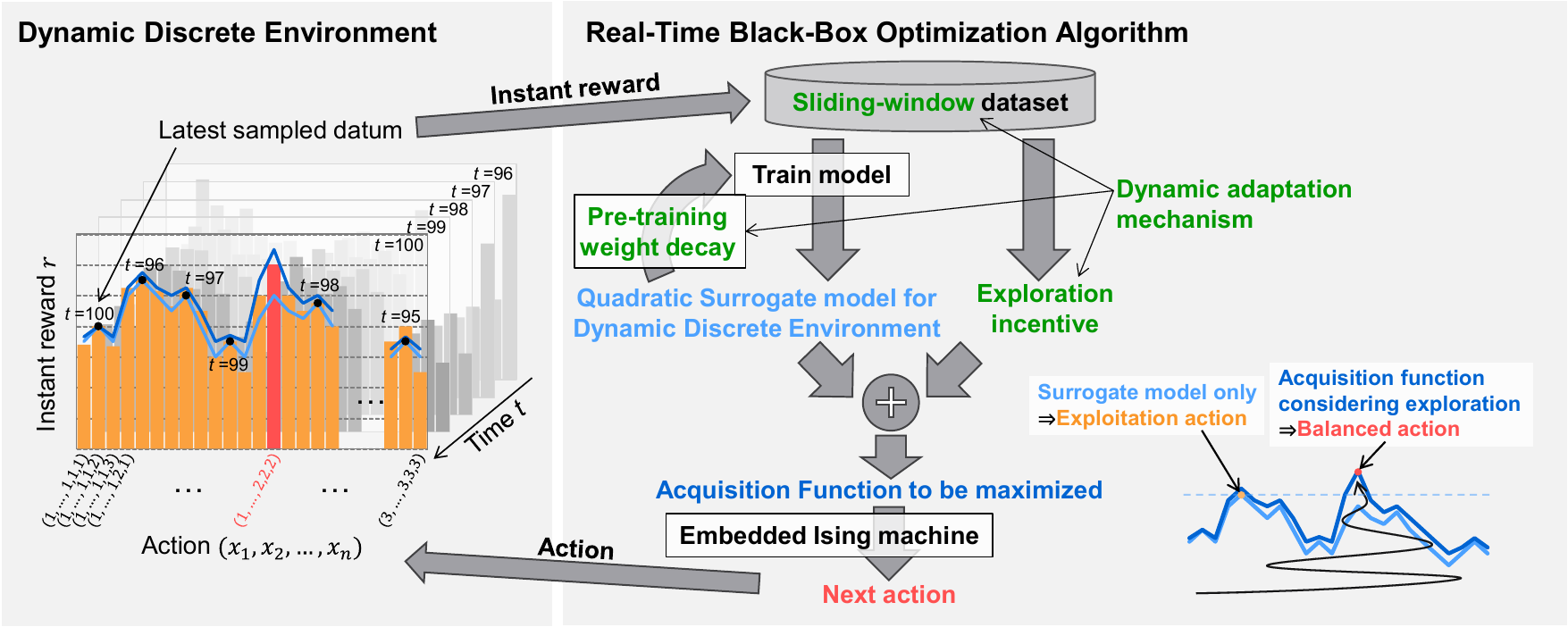}
\caption{
Dynamic discrete environment and Real-Time Black-Box Optimization Algorithm:
The dynamic adaptation mechanism enables the quadratic surrogate model to adapt to changes in the discrete environment while also adjusting the exploration-exploitation balance.
The embedded Ising machine quickly finds the next action as an approximate solution to maximize the acquisition function that is the sum of the surrogate model and an incentive function for enhancing the exploration.
}
\label{Fig_RTBBO}
\end{figure}

Here, we propose and demonstrate a Real-Time Black-Box Optimization (RT-BBO), a heuristic algorithm designed for dynamic MAB problems where actions are represented by many interacting discrete variables (Fig. \ref{Fig_RTBBO}).
``Black-Box Optimization'' refers to the high-dimensional optimization capability derived from FMQA, while “Real-Time” indicates the method’s adaptability to dynamic environmental changes and its suitability for embedding into practical real-time systems.
The algorithm assumes that the number of possible actions is too large for conventional MAB algorithms and that the changes in the environment are gradual enough that a machine learning method can work effectively.
The RT-BBO incorporates a dynamic adaptation mechanism into the scheme of the FMQA and utilizes an embedded Ising machine instead of an external Ising machine.

The dynamic adaptation mechanism enables the surrogate model to adapt to dynamic changes in the environment.
This mechanism consists of three approaches.
The first approach is a sliding-window dataset, which restricts the dataset used for machine learning to recent data.
The second approach is pre-training weight decay, which makes the existing model parameters slightly smaller before training the model each time, similar to weight decay in machine learning \cite{loshchilov17} but being different in execution timing.
The third approach is an exploration incentive, which encourages the Ising machine to select a balanced action that considers both the exploration and exploitation rather than an exploitation-biased action that maximizes the surrogate model as in the FMQA.
In addition, for the case that an instant reward comprises multiple instant rewards (observable sub-rewards in a sampling cycle), we show a technique that enhances the adaptability to the changes in environment.

The embedded Ising machine broadens the applicability of Ising machines to edge and embedded systems.
Instead of the external Ising machine accessed via the internet as seen in the FMQA~\cite{kitai20}, we employ a Simulated Bifurcation (SB)-based Ising machine \cite{sbm1, sbm2} as the embedded Ising machine. The SB algorithm, which was derived from the classical counterpart of a quantum computer named Quantum bifurcation Machine (QbM) \cite{qbm}, has a high degree of parallelism and enables high-speed implementations using parallel computers~\cite{FPL19, NatEle, Kashimata24}.
The SB-based embedded Ising machine can find good solutions in a single-shot execution and is fast enough (low latency) to be used for real-time financial trading systems~\cite{ISCAS20, ACCESS23a, ACCESS23b}, in-vehicle multiple object tracking systems~\cite{MOT24}, and an iterative optimization algorithm~\cite{Matsumoto22}.

We prototype a wireless control system to demonstrate the dynamic adaptability of the proposed RT-BBO method for practical environments.
The assumed wireless cellular network (environment) includes multiple base stations that simultaneously use the same frequency to communicate with their respective users to improve the communication throughput with limited bandwidth.
The primary issue in the wireless environment is throughput degradation due to interference between neighboring cells.
The control system adjusts beamforming patterns that determine the beam direction and transmission power for each base station (actions represented with interacting discrete variables) and aims to increase the average throughput (average rewards) of the entire network by reducing the interferences. 
This problem, known as the ``beamforming coordination problem,'' has been tackled using deep reinforcement learning \cite{Mismar20,DRL_for_DDBC,Zhou23} that requires observations of internal states of the environment (environmental observations) other than the rewards.
The proposed method optimizes the beamforming patterns based solely on the achieved throughputs of all the base stations (multiple instant rewards) in spite of the dynamic environment changes due to moving users and the effects of Rayleigh fading.

\section{Results}\label{sec_results}
\subsection{Algorithm} \label{subsec_algorithm}

\begin{figure}[tb]
\centering
\includegraphics[width=17.2 cm]{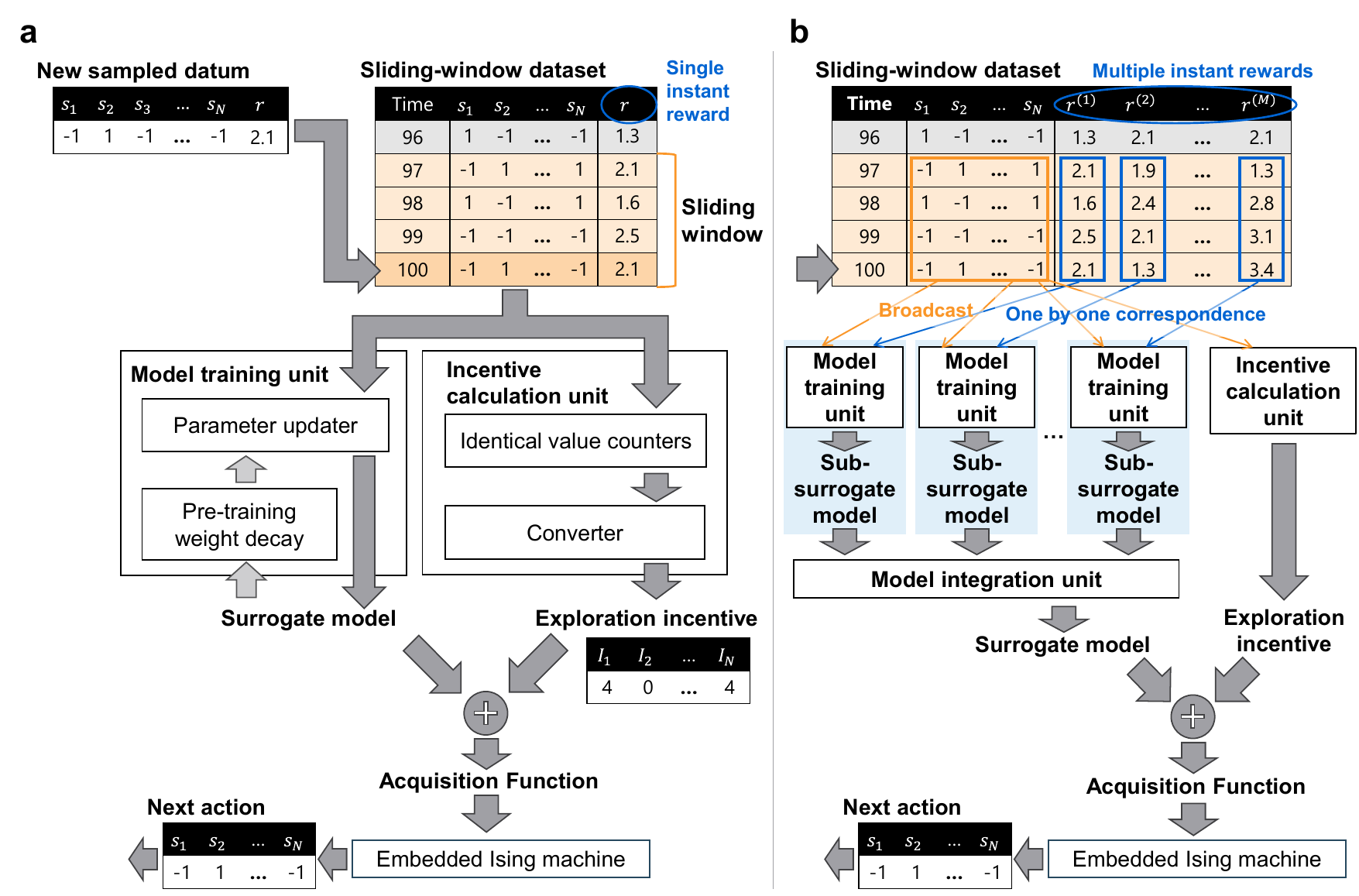}
\caption{Dataflow graph of the RT-BBO algorithm:
(a) RT-BBO for SR (single instant reward):
Pre-training weight decay is applied to the surrogate model, and then the model is trained to approximate the environment using the sliding-window dataset.
The exploration incentive for each spin variable is the square of the number of successive identical values for each spin variable in the sampled data (elapsed time since the last change of the spin value).
The embedded Ising machine finds the next action that maximizes the acquisition function, which is the sum of the surrogate model, the incentive, and the penalty terms for one-hot encoding.
(b) RT-BBO for MR (multiple instant rewards):
Sub-surrogate models corresponding to each reward are trained independently and then are integrated into one surrogate model.} 
\label{Fig_describe}
\end{figure}

The main feature of the RT-BBO is its adaptability to dynamically changing environments.
The FMQA~\cite{kitai20}, which forms the basis of the RT-BBO, is designed to find the semi-optimal actions in static environments.
The basic FMQA algorithm builds a quadratic surrogate model using observed results, optimizes the variables in the model with an Ising machine, and selects the next action based on the optimized results.
To effectively optimize actions in dynamically changing environments, the surrogate model must adapt rapidly to the current environment.
Therefore, the core focus of the RT-BBO lies in the adaptability of its surrogate model to such environments.
The proposed RT-BBO heuristically addresses the problem of maximizing total rewards by taking actions and receiving rewards in a dynamically changing environment that requires combinatorial optimization.
An example of such a problem is beamforming coordination in wireless cellular networks, where the goal is to determine the optimal beam directions and the transmission power levels of base stations in highly interfering environments (as discussed in Sec. 2.2).

Fig.\ref{Fig_describe}\textbf{a} illustrates the dataflow of the RT-BBO. In each sampling cycle, it takes as input a newly sampled datum, comprising the current action and its corresponding instant reward, and outputs the next action. As shown in Fig.\ref{Fig_RTBBO}, an instant reward is represented as a real number ($r$), and an action in a discrete environment with $n$ control inputs is represented as a vector of discrete values ($x_1, ..., x_n$), where each control input $x_l$ takes one of $d_l$ possible states. This vector ($x_1, ..., x_n$) is encoded into, and decoded from, a vector of $N$ spin variables ($s_1, ..., s_N$), where $N = \sum_l d_l$ and $s_i \in {-1, 1}$, using an encoding/decoding method described in the Methods section. A constraint imposed by this encoding/decoding process is incorporated into the acquisition function as a penalty term, $H_\mathrm{encoding}(\mathvect{s})$.

\subsubsection{Surrogate Model}

To represent surrogate models, both the FMQA and the RT-BBO employ a technique known as the factorization machine~\cite{rendle10}, which enables low-rank approximation to prevent overfitting and accelerates training by reducing the number of parameters. The surrogate model $\hat{r}(\mathvect{s})$ is defined as:

\begin{equation}\label{Eq_yhat}
\hat{r}(\mathvect{s}) = \sum_{i=1}^{N} \sum_{j=i+1}^{N} \langle \mathvect{v}_i, \mathvect{v}_j \rangle s_i s_j + \sum_{i=1}^N w_i s_i + w_0,
\end{equation}
where $\mathvect{v}_i$, $\mathvect{v}_j$, and $w_i$ are model parameters, and $s_i$ are spin variables.
These parameters are updated through machine learning but are treated as fixed coefficients when passed to the Ising machine. Conversely, the spin variables are optimized by the Ising machine but serve as feature variables during model training. The factorization machine expresses each quadratic coefficient as the inner product of two vectors ($\mathvect{v}_i$ and $\mathvect{v}_j$) associated with the spin variables ($s_i$ and $s_j$). The dimension of $\mathvect{v}_i$ and $\mathvect{v}_j$ is $K$, which is smaller than $N$ (low-rank approximation).

The model is trained using spin-variable representations of actions from the dataset as explanatory variables, and instant rewards as the target variable. The parameter updater in Fig.\ref{Fig_describe}\textbf{a} adjusts the parameters ($\mathvect{v}_i$ and $w_i$) to minimize prediction error using a backpropagation method based on the partial derivatives of the loss function. To enhance robustness against noisy data, the updater employs the log-cosh loss function\cite{saleh2024} and a mini-batch learning strategy. For further details on the factorization machine and the machine learning process, refer to the Methods section.

\subsubsection{Sliding Window}
When training the surrogate model, the key mechanism that enables adaptation in the RT-BBO is the use of a sliding-window dataset.

In the FMQA for static environments (i.e., a static BBO), the surrogate model approximating the environment is trained using a cumulative dataset that records all historical pairs of actions and instant rewards. In contrast, the RT-BBO (Fig.~\ref{Fig_RTBBO}) for dynamic environments constructs its surrogate model using a sliding-window dataset, which retains only the most recent data within a fixed window size, discarding older data as new samples arrive.

For comparison, we introduce a simple FMSB (a baseline), which combines the factorization machine with the SB-based Ising machine without any dynamic adaptation mechanism, as a baseline method.
The surrogate model for the baseline is trained using randomly selected data from the cumulative dataset, ensuring that the number of training samples per cycle matches that of the RT-BBO to equalize computational cost.
Note that the original FMQA does not use random sampling but instead trains on the full historical dataset at each training.

\begin{figure}[tb]
\centering
\includegraphics[width=17.2 cm]{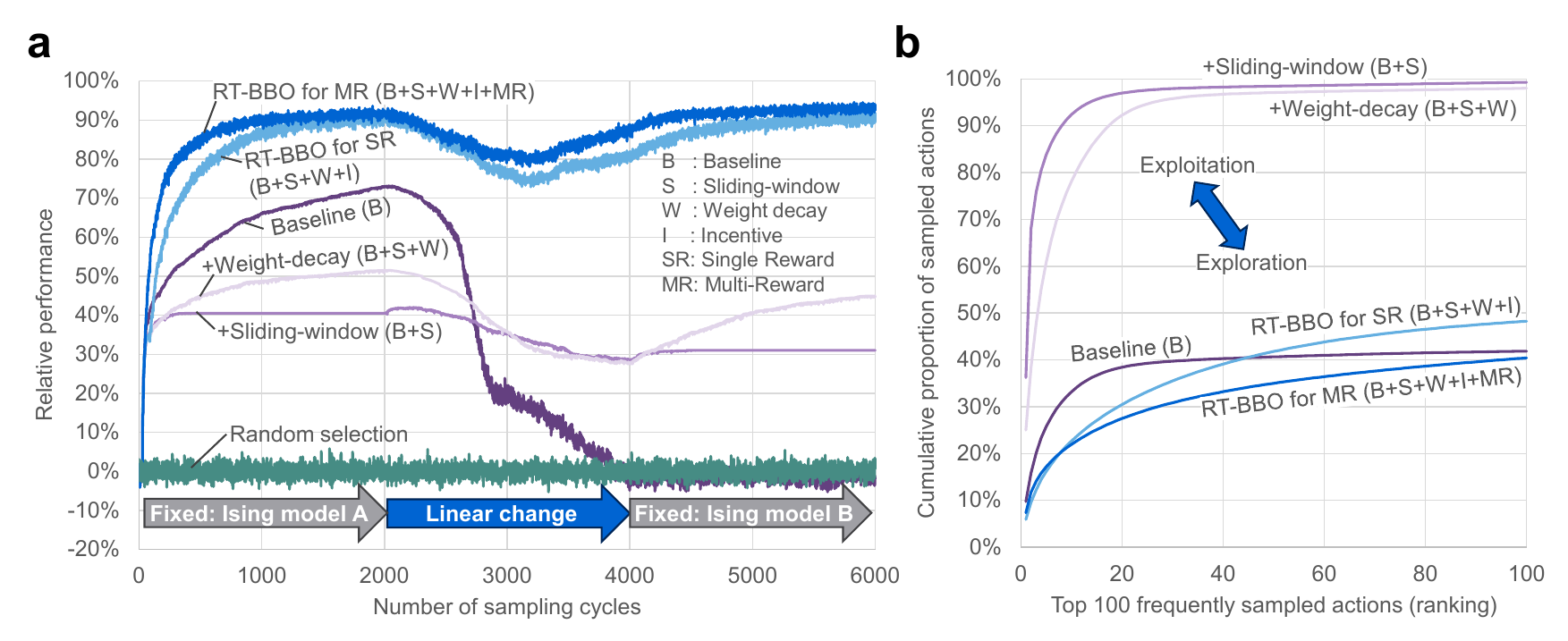}
\caption{Adaptability of the RT-BBO algorithms (i.e., FMSB with dynamic adaptation mechanism) in a dynamically changing environment.
(a) Comparison of optimization performance when techniques that constitute the dynamic adaptation mechanism (S: sliding-window, W: weight decay, I: incentive, SR: single-reward, MR: multi-reward) are added one by one to the baseline (i.e., FMSB without dynamic adaptation mechanism). A black-box function is fixed to $M_\text{A}(\mathvect{s})$ during 0--2000 sampling cycles and to $M_\text{B}(\mathvect{s})$ during 4000--6000 sampling cycles, changing in 2000--4000 sampling cycles as expressed by $\alpha M_\text{A}(\mathvect{s})+(1-\alpha)M_\text{B}(\mathvect{s})$. $M_\text{A}$ and $M_\text{B}$ consist of ten randomly generated Ising models, where the Ising energy of each model corresponds to one of multiple instant rewards in Fig.~\ref{Fig_describe}, and the sum of ten Ising energies corresponds to the single (total) reward in Fig.~\ref{Fig_describe}. The expected value of the total reward for a randomly selected solution (random-selection) is zero, and the total rewards observed for the comparable techniques are normalized to the total reward obtained by the SB-based Ising machine when solving the black-box function as a white-box. The vertical axis (relative performance) corresponds to the averages of the normalized total rewards over 50 independent trials for the same black-box function. (b) Cumulative proportion (ratios) of the top 100 frequently sampled actions out of 6000 actions selected in (a). If a line passes through the lower right (/upper left), it indicates a preference for exploration (/exploitation).
}
\label{Fig_randomising}
\end{figure}

Fig.~\ref{Fig_randomising}\textbf{a} shows the enhancement in dynamic adaptability when techniques that constitute the dynamic adaptation mechanism are added one by one to the baseline, where a black-box function changes during the 2000--4000 sampling cycles. The baseline lacks the capability to adapt to environmental changes, as depicted by its behavior during the 2000--4000 cycles. In comparison, the sliding-window technique provides a certain level of adaptation even in the dynamically changing environment. However, the optimization accuracy remains lower than the ideal scenario where the black-box is treated as a white-box.

This issue arises from poor exploration. Fig.~\ref{Fig_randomising}\textbf{b} depicts the cumulative proportions of the top 100 frequently sampled actions out of 6000 actions selected in Fig.~\ref{Fig_randomising}\textbf{a}. When using the sliding-window technique alone, the system tends to sample a limited set of actions repeatedly. Once the surrogate model captures the characteristics of a local solution, it becomes biased toward sampling actions near that solution, thereby missing opportunities to explore better alternatives. To address this issue, we introduce pre-training weight decay and an exploration incentive, as described in the following subsections.

\subsubsection{Pre-training Weight Decay}
To encourage exploration and enhance adaptability, it is effective to attenuate the influence of previously learned surrogate models at each sampling cycle. The RT-BBO framework repeats sampling iterations (hereafter, outer loop), which include training iterations for updating the surrogate model (hereafter, inner loop). Refer to Algorithm 1 in the Methods section for more details. The pre-training weight decay, applied before the training iteration (the inner loop) in every sampling cycle (once per outer loop), scales the model parameters by a coefficient slightly less than one ($c_\mathrm{decay}$). This causes the surrogate model to temporarily diverge from the environment, allowing recent data to have a stronger influence during subsequent training iterations.

When training a regression model, weight decay is conventionally performed inside the training iteration (in the inner loop) to prevent overfitting as a form of regularization. In contrast, the proposed Pre-training Weight Decay is performed once before the training iteration to intentionally induce a relatively large deviation of the model to emphasize the recent data.

\subsubsection{Exploration Incentive}
To encourage exploration, the RT-BBO introduces an incentive mechanism that promotes diverse sampling. In the FMQA framework~\cite{kitai20}, the Ising machine is used to find the minimizer (or maximizer) of the surrogate model, focusing solely on exploitation. In dynamic environments, where the optimal action may change over time, continuous and effective exploration is necessary to adapt to these changes. The proposed incentive mechanism encourages changes in spin variables based on the elapsed time since their last change, helping to avoid getting trapped in local optima.

The incentive, $H_\mathrm{exploration}(\mathvect{s})$, is defined as:

\begin{equation}\label{Eq_incentive}
H_\mathrm{exploration}(\mathvect{s}) = - c_\mathrm{exploration} \sum_i I_i s_i,
\end{equation}
where $I_i$ is the incentive intensity for each spin variable ($s_i$), and $c_\mathrm{exploration}$ controls the overall intensity of the incentive.

The incentive calculation unit in Fig.~\ref{Fig_describe}\textbf{a} computes the intensity ($I_i$) using counters that track how many consecutive times each spin variable has taken the same value repeatedly. 
Each counter increments if the current spin value matches the previous one, and resets to zero otherwise.
The converter in the unit squares each counter value to emphasize longer repetitions, then multiplies each squared value by the corresponding last spin value ($\{-1, +1\}$) to determine the sign of ($I_i$).

To keep the average number of consecutive identical spin values (i.e., counter values) within a desirable range (a hyperparameter), $c_\mathrm{exploration}$ is dynamically adjusted: it increases when the average exceeds the upper bound of the range and decreases when it falls below the lower bound.
This mechanism helps maintain the average elapsed time within the target range, thereby improving robustness to variations in the dynamic range of the surrogate model parameters.

\subsubsection{Acquisition Function}
The acquisition function in this work (see Fig.~\ref{Fig_RTBBO}) is the $\operatorname{argmax}$ of the sum of the updated surrogate model, the incentive, and the encoding penalty. The embedded Ising machine finds a spin configuration ($\mathvect{s}$) that maximizes the acquisition function, and the spin configuration found is the next action. This optimization problem is expressed by:

\begin{equation}\label{Eq_acq2}
\mathvect{s} \simeq \underset{\mathvect{s} \in \{-1, +1\}^N} {\operatorname{argmax}} \left( \hat{r}(\mathvect{s}) +  H_\mathrm{exploration}(\mathvect{s}) + H_\mathrm{encoding}(\mathvect{s}) \right).
\end{equation}

As the embedded Ising machine, we use an SB-based Ising machine, which is particularly suitable for the RT-BBO due to its high speed, making it appropriate for real-time applications~\cite{ISCAS20, ACCESS23a, ACCESS23b, MOT24, Matsumoto22}. The SB algorithm offers greater parallelism than simulated annealing-based methods and can be accelerated using parallel computing~\cite{FPL19, NatEle}. Note that while Eq.~\ref{Eq_acq2} is formulated as a maximization problem, Ising machines are typically designed for minimization. To align with the actual RT-BBO implementations, the $\operatorname{argmax}$ in Eq.~\ref{Eq_acq2} is converted to $\operatorname{argmin}$, and the signs of some terms are adjusted accordingly.

In Figures~\ref{Fig_describe}\textbf{a} and \ref{Fig_randomising}, the FMSB enhanced with the sliding-window dataset, the pre-training weight decay, and the incentive is referred to as RT-BBO for SR (single instant reward). The performance of the RT-BBO can be further improved in special cases where the total instant reward is the sum (or a linear combination) of multiple observable instant rewards. Fig.~\ref{Fig_describe}\textbf{b} illustrates the dataflow of RT-BBO for MR (multiple instant rewards), where multiple rewards are stored in the dataset and used to independently train sub-surrogate models. These sub-models are, after training, combined into an integrated surrogate model as:

\begin{equation}\label{Eq_yhat_div}
\hat{r} = \sum_m p_m \hat{r}^{(m)},
\end{equation}
where $\hat{r}^{(m)}$ is the $m$-th sub-surrogate model and $p_m$ is its weight (typically set to 1). 
Other than this, the process remains the same as in the single-reward case.
This approach decomposes a large approximation problem into several smaller ones, improving the overall approximation quality, as shown in Figs.~\ref{Fig_randomising} and~\ref{Fig_result}.

As shown in Fig.~\ref{Fig_randomising}\textbf{a}, introducing weight decay and the incentive to the baseline with a sliding window significantly improves both optimization performance and adaptability.
Furthermore, the multi-reward modeling enhances these improvements even further.
Compared to the Ising energy achieved by the SB-based Ising machine when solving the Ising model as a white-box, the RT-BBO for MR achieves approximately 90\% relative performance in static environments and maintains around 80\% in dynamic environments by repeatedly taking actions and observing rewards, without accessing the inside of the black box.

\subsection{Demonstration}
We demonstrate the adaptability of the RT-BBO algorithms in optimizing wireless communication by developing an RT-BBO-based beamforming optimizer that interacts with a simulator of a dynamically changing wireless environment. This simulator simulates the internal states of a cellular network, consisting of base stations and mobile users. The actions, represented by discrete variables and determined by the RT-BBO algorithms, correspond to the control operations of the base stations. After each action, an instant reward, which is defined as the total communication throughput between the base stations and users, is observed. The function that describes the instant reward based on the action is highly complex and is far from the form of a quadratic discrete function. In the RT-BBO, this function (the black-box function) is approximated using a quadratic surrogate model. 

\subsubsection{Wireless Environment}
\begin{figure}[tb]
\centering
\includegraphics[width=17.2 cm]{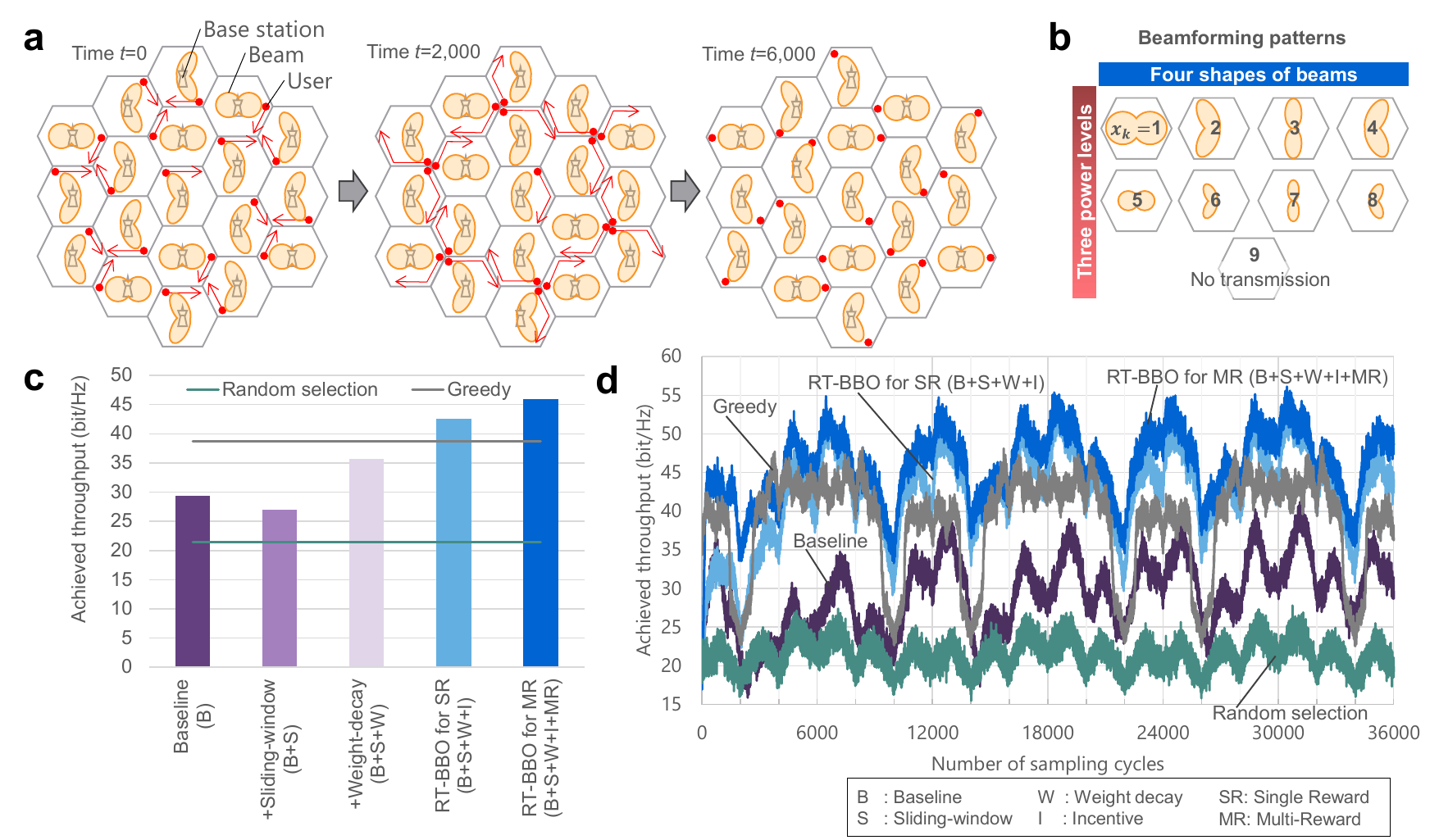}
\caption{Experimental results on the prototype for the wireless control system.
(a) Snapshots showing the situation of moving users and base stations: The users move along the edges of the cells, gradually gathering into groups of three users (t=0 $\rightarrow$ 2000), and then moving apart (t=2000 $\rightarrow$ 6000).
(b) The beamforming patterns for each base station: Four shapes and three transmission power levels and thus there are nine patterns in total.
(c) Performance improvement with dynamic adaptation mechanism: There is little difference in performance between the baseline method (FMSB without dynamic adaptation mechanism) and the case where only the sliding-window dataset is introduced. Performance improves with the use of weight decay, incentive, and multi-reward modeling.
(d) Achieved throughput in the environment (instant reward): The total achieved throughputs of all the base stations using the RT-BBO for MR (multiple instant rewards) and for SR (single instant reward), which are compared to the baseline, a random selection method, and a greedy method that requires environmental observations (internal information in the black box function). This result is the average of 50 independent trials. See the Supplementary video for more details.
}
\label{Fig_result}
\end{figure}

Fig.~\ref{Fig_result}\textbf{a} illustrates a small-cell network in which multiple base stations are located at the centers of cells and are assumed to operate on the same communication frequency to maximize throughput under limited bandwidth.
Signals from the base stations interfere with one another.
This interference is mitigated using a beamforming technique, which allows each base station to direct radio waves in specific directions by adjusting the amplitude and phase of signals transmitted from a set of antenna elements on a base station.
Note that only the one-way transmission of radio waves from the base stations to the users is considered in this demonstration.

As shown in Fig.~\ref{Fig_result}\textbf{b}, each base station has nine predefined beamforming patterns, consisting of four directional shapes (a codebook) and three transmission power levels (maximum, medium, and no transmission). The variable $x_i$ denotes the index of the beamforming pattern selected for the $i$-th base station. The set of beamforming patterns for all base stations constitutes the action for RT-BBO, represented by $n$ discrete variables $(x_1, ..., x_n)$, where $n$ is the number of base stations. In our experiment, $n = 19$, resulting in a total of $9^{19} \approx 1.35 \times 10^{18}$ possible actions. As we use nine spin variables for each $x_i$ (see the subsection entitled one-hot encoding in the Method section), the total spin variable needed is 171.

To make the demonstration results easier to interpret, we assume the following conditions for moving users. The number of users is equal to $n$, and each base station communicates with one user located within its cell. The base station–user pairs are predetermined, and each pair may experience interference from all other base stations. The users continuously move along the edges of the cells. As shown in Fig.~\ref{Fig_result}\textbf{a}, from time $t = 0$ to $t = 2{,}000$, groups of three adjacent users gather at the intersection of their respective cell areas. At $t=2{,}000$, the users' positions lead to severe interference among base stations, making optimization necessary to maintain high throughput. From $t = 2{,}000$ to $t = 6{,}000$, the users move apart again. From $t = 6{,}000$ to $t = 36{,}000$, the users repeatedly retrace their paths from $t = 0$ to $t = 6{,}000$.

The communication throughput for each base station–user pair is calculated using a wireless communication model, consistent with the one used in Ref.~\cite{DRL_for_DDBC}. In this model, the amplitude and phase of the signal transmitted by the $i$-th base station with beamforming pattern $x_i$ are represented by a complex vector $\mathvect{w}_i(x_i)$. The channel vector between the $i$-th base station and the $j$-th user, which captures propagation effects including Rayleigh fading, is denoted by $\mathvect{h}_{i,j}$. The received signal at the $j$-th user from the $i$-th base station is given by $\mathvect{h}^{\dagger}_{i,j} \mathvect{w}_i(x_i)$.

The throughput of the $k$-th base station, $C_k$, is calculated as:

\begin{equation}\label{Eq_throughput}
C_k(\mathvect{x}) = \log_2 (1+\psi_k(\mathvect{x})),
\end{equation}
where $\psi_k$ is the signal-to-interference-plus-noise ratio (SINR) for the $k$-th base station–user pair, and $\mathvect{x} = (x_1, ..., x_n)^T$ represents the beamforming pattern vector. The SINR is computed as:

\begin{equation}\label{Eq_SINR}
\psi_k(\mathvect{x}) = \cfrac{\left| \mathvect{h}^{\dagger}_{k,k} \mathvect{w}_k(x_k) \right|^2}{\sum_{j \ne k} \left| \mathvect{h}^{\dagger}_{j,k} \mathvect{w}_j(x_j) \right|^2 + \sigma^2},
\end{equation}
which is the ratio of the received signal power from the intended base station to the sum of interference from other base stations and background noise.

The total throughput, $\sum_k C_k(\mathvect{x})$, serves as the total instant reward (or the sum of multiple instant rewards) corresponding to the beamforming pattern vector $\mathvect{x}$, which is treated as the action in the RT-BBO.

The channel vector $\mathvect{h}_{i,j}$ accounts for both the effects of user mobility and Rayleigh fading, the latter of which models fluctuations in signal amplitude and phase caused by multipath propagation. For further details on the wireless simulator settings, refer to the Methods section.

\subsubsection{Evaluation}
In this evaluation, we compare the average rewards (i.e., achieved throughputs) of FMSB-based beamforming optimizers, including the proposed RT-BBO, against a greedy method and a random selection method.

The greedy method selects the beamforming pattern that maximizes the received power for the target user, based on the user's location (internal information of the black-box function). Specifically, the received power, $\left| \mathvect{h}^{\dagger}_{k,k} \mathvect{w}_k(x_k) \right|^2$ is evaluated for each candidate beamforming pattern $x_k$, and the pattern yielding the highest value is selected. This procedure corresponds to maximizing only the numerator of Eq. 6, which represents the received signal power from the corresponding base station, while ignoring the denominator that accounts for inter-cell interference. Consequently, the method yields an optimal solution under the assumption of no interference from other base stations.
The random selection method uniformly chooses beamforming patterns for each base station from the nine predefined options, without using any specific information.
We evaluated five versions of FMSBs, starting from a baseline and incrementally adding the following techniques: sliding-window dataset, pre-training weight decay, exploration incentive, and multi-reward modeling.

Fig.~\ref{Fig_result}\textbf{c} presents a comparison of the average throughputs. All FMSB variants outperform the random selection method. Both the RT-BBO for SR and for MR achieve higher throughput than the greedy method. The baseline and the version with only the sliding-window dataset show similar performance. The effectiveness of weight decay, incentive, and multi-reward modeling is confirmed, as performance improves steadily with each addition.

Fig.~\ref{Fig_result}\textbf{d} shows the throughput transition over sampling cycles (time). Under severe interference conditions, such as at $t = 2{,}000$, the throughput of the greedy method drops significantly. In contrast, the RT-BBO implementations maintain higher throughput, achieving better optimization despite having less information. As also observed in Fig.~\ref{Fig_randomising}\textbf{a}, the RT-BBO for MR outperforms the RT-BBO for SR, particularly in the early phase, and maintains a slight advantage throughout the evaluation.

\section{Discussion}\label{sec_discussion}
We have proposed an RT-BBO method, a heuristic MAB algorithm, for dynamic environments where each action is represented by many interacting discrete variables. The RT-BBO incorporates a dynamic adaptation mechanism into an existing discrete BBO algorithm (FMQA~\cite{kitai20}) for static environments. This dynamic adaptation mechanism consists of a sliding-window dataset, pre-training weight decay, and exploration incentive, enabling the quadratic surrogate model to adapt to changes in the discrete environment while also adjusting the exploration-exploitation balance. The RT-BBO relies on an SB-based low-latency embedded Ising machine to find the next action that maximizes an acquisition function, obtained by combining the surrogate model and the incentive, from the enormous number of possible actions. The prototyped wireless control system for a cellular network has demonstrated the dynamic adaptability of the proposed method in a real-time application.

There are two possible directions for future work.
The first direction is to apply the RT-BBO to various real-time systems to clarify its applicability; for example, recommendation systems for advertisements \cite{schwartz2017} and dynamic pricing systems for online shops \cite{den2015}.
The second direction is to enhance the RT-BBO to consider context; such enhanced algorithms may have use cases in personalized recommendations \cite{li2010} and stock investments \cite{cannelli2023}.

\clearpage

\section*{Methods} \label{sec_methods}
\subsection*{Ising Model and Ising machine}
Many combinatorial optimization problems can be reduced to the energy minimization problem of the Ising model \cite{Lucas}. The energy in the Ising model is expressed as 

\begin{equation}\label{Eq_energy}
H = -\cfrac{1}{2} \sum_{i=1}^{N} \sum_{j=1}^{N} J_{i,j} s_i s_j + \sum_{i=1}^{N} h_i s_i.
\end{equation}
Here, the $N$ variables $s_i$, called spin variables, take binary values of +1 and -1.
A computer specialized in calculating the spin configuration that minimize this energy is called an Ising machine. Ising machines aim to quickly find approximate solutions. Note that calculating the exact solution in a practical (polynomial) time is still not guaranteed with current technology~\cite{complexity}.
The algorithms of Ising machines are classified as quantum algorithms \cite{D-wave} and classical algorithms \cite{ising_compare, DA, STATICA, amorphica, honjo21,kalinin20,PoorCIM19,moy22,albertsson21,wang21,SimCIM21,Graber24}.
The Ising machines based on quantum algorithms\cite{D-wave} are difficult to be embedded in real-time systems because they require special equipment like refrigerators. Though some Ising machines based on classical algorithms, such as simulated annealing-based Ising machines \cite{DA, STATICA, amorphica}, have opportunities to be embedded in real-time systems, there are few studies for embedded Ising machines.
We have proposed use cases for SB-based embedded Ising machines: real-time financial trading systems~\cite{ISCAS20, ACCESS23a, ACCESS23b}, in-vehicle multiple object tracking systems~\cite{MOT24}, and an iterative optimization algorithm~\cite{Matsumoto22}.

Simulated Bifurcation (SB) algorithms are quantum-inspired classical oscillator-based algorithms.
The SB algorithm simulates adiabatic time evolutions of nonlinear Hamiltonian systems having oscillators corresponding to spins in the Ising model.
Its Hamiltonian is designed to exhibit bifurcation phenomena.
The algorithm used in the RT-BBO is ballistic SB~\cite{sbm2}, which performs well at calculating approximate solutions in a short time~\cite{ISCAS20, ACCESS23a, ACCESS23b,MOT24,Matsumoto22}.
The positions $x_i$ and momenta $y_i$ of oscillators corresponding to the spin variables are updated according to the following equations.

\begin{align}\label{Eq_sbm}
y^{t_{k+1}}_{i} &\leftarrow y_{i}^{t_k} + \left[ - \left( a_0 - a^{t_k} \right) x_{i}^{t_k} - \eta h_i + c_0 \sum_{j}^{N} J_{i,j} x_{j}^{t_k} \right] \Delta_t,\\
x_{i}^{t_{k+1}} &\leftarrow x_{i}^{t_k} + a_0 y_i^{t_{k+1}} \Delta_t, \\
(x_{i}^{t_{k+1}}, y_{i}^{t_{k+1}}) & \leftarrow 
\begin{cases}
(\text{sgn}(x_{i}^{t_{k+1}}),0) & \text{if $| x_{i}^{t_{k+1}} | > 1$}, \\
(x_{i}^{t_{k+1}}, y_{i}^{t_{k+1}}) & \text{if $| x_{i}^{t_{k+1}} | \le 1$},
\end{cases}
\end{align}
where $a_0$, $c_0$, $\eta$, and $\Delta_t$ are hyper-parameters, $\Delta_t$ is the time step of simulation of SB, and $\text{sgn}(x)$ is the sign function. The parameter $a^{t_k}$ evolves linearly from 0 to 1.
The SB-based Ising machine repeats this update procedure for the predetermined number of times (1,000 in this work).
The output spin values are the binarized results of the final positions.

\subsection*{Factorization Machine}
Factorization Machines (FMs) are models designed for supervised learning tasks. They effectively capture feature interactions within a feature vector using a low-rank approximation~\cite{rendle10}. While general FMs are capable of modeling arbitrary-order interactions among features, our focus is specifically on second-order interactions, which are central to our work.

To efficiently represent these second-order interactions, FMs approximate a square matrix in $\mathbb{R}^{N \times N}$ using a low-rank decomposition. This is achieved through a matrix of factor vectors, denoted as $V = (v_{ik})$, where $V \in \mathbb{R}^{N \times K}$. The compressed representation is then expanded into a square matrix $J$ via the product $J = VV^\mathsf{T}$.
Each element of $J$ is computed as:

\begin{equation}\label{Eq_fm}
J_{ij} = \langle \mathvect{v}_i , \mathvect{v}_j \rangle = \sum_{k=1}^{K} v_{ik} v_{jk},
\end{equation}
where $\mathvect{v}_i$ is the factor vector corresponding to the $i$-th feature, and $v_{ik}$ is its $k$-th component. 
In other words, each pairwise interaction $J_{ij}$ is approximated by the similarity between the $i$-th and $j$-th feature vectors.

The dimensionality $K$ of the factor vectors determines the rank of the reconstructed matrix $J$ and is typically much smaller than $N$ (e.g., $K = 6$ in this work). 
This low-rank factorization reduces the number of parameters required to model quadratic interactions from $(N^2 - N)/2$ to $NK$. 
Such compression not only accelerates the learning process but also mitigates overfitting, which is an important consideration when working with limited data.

\subsection*{One-hot encoding}
Before and after the core RT-BBO algorithm, one-hot encoding and decoding are applied to extend the RT-BBO to handle discrete variables that take more than two values, since Ising machines take spin variables that take either one of $+1$ or $-1$.
The one-hot encoding converts $n$ discrete values ($x_1, ..., x_n$) representing the taken action into $N$ spin values ($s_1, ..., s_N$) before adding them to the dataset, and reversely converts the spin values output from the Ising machine to discrete values in the original format representing the next action.
Here, we assume $l$-th discrete variable ($x_l$) has $d_l$ possible values.
By the encoding, the $l$-th discrete variable ($x_l$) is represented with a set of $d_l$ spin variables.
Within each set, only one spin variable corresponding to the value of $x_l$ is $+1$ at a time, while the other variables are set to be $-1$. Thus, $N$ spin variables are used to represent $n$ discrete values ($N = \sum_{l=1}^n d_l$).
To enforce this, RT-BBO adds penalty terms to the acquisition function as \cite{seki22} did.
The penalty term is described as

\begin{equation}\label{Eq_encoding}
H_\mathrm{encoding}(\mathvect{s}) = - \cfrac{c_\mathrm{encoding}}{4} \sum_{l=1}^{n} \left( \left( \sum_i ^{d_{l}} s_{d_l\cdot i+l} \right) + d_{l} - 2 \right)^2,
\end{equation}
where $c_\mathrm{encoding}$ is a coefficient to control the strength of the constraints.
High $c_\mathrm{encoding}$ prevents the Ising machine from outputting solutions that violate the constraints.
The coefficient $c_\mathrm{encoding}$ should be adjusted because too high or too low a coefficient degrades the quality of solution output from Ising machines.

The proposed exploration incentive can work with the one-hot encoding;
in the scheme of the one-hot encoding, the incentive encourages the Ising machine to explore values that have not been selected recently.

\subsection*{Model training unit}
The model training unit updates the parameters of the surrogate model based on the data in the sliding-window dataset.
Algorithm~\ref{algo_mtu} outlines the procedure used in the model training unit, which consists of two main parts: pre-training weight decay and a parameter updater.
The parameter updater is based on the original FMQA, with several modifications to improve computational efficiency and robustness against noisy data.

First, we adopted the mini-batch method, which uses only a subset of the extracted data in the dataset for gradient calculation (20 data points for our implementations), instead of using all the data in the dataset to reduce computational cost.
The number of extracted data (mini-batch size) should be experimentally adjusted because if it is too large, the computational cost increases, and if it is too small, the stability of the gradient is lost. For example, if the noise level in the instant reward is high or the black-box function is very complex, using more data points can help stabilize the gradient~\cite{rolnick2018}.
During each sampling cycle, the gradient update is repeated $N_\mathrm{train}$ times (200 for our implementations). While a larger $N_\mathrm{train}$ increases computational cost, it can also lead to better performance and should be chosen based on the trade-off between sampling frequency and optimization quality.

Second, the parameter updater employed the log-cosh loss function~\cite{saleh2024}, defined as

\begin{equation}\label{Eq_loss}
L(r(\mathvect{s}), \hat{r}(\mathvect{s})) = \log(\cosh(\hat{r}(\mathvect{s}) - r(\mathvect{s}))),
\end{equation}
where $r(\mathvect{s})$ is the observed reward and $\hat{r}(\mathvect{s})$ is the model prediction.
Compared to the mean squared error used in the original FMQA, the log-cosh loss is more robust to outliers~\cite{saleh2024}.
The gradients of the loss with respect to the model parameters are given by
\begin{align}\label{Eq_loss_partial}
\cfrac{\partial L}{\partial w_i}   &= \cfrac{\partial L}{\partial \hat{r}} \cdot \cfrac{\partial \hat{r}}{\partial w_i} =  \tanh(\hat{r}(\mathvect{s}) - r(\mathvect{s})) \cdot s_i, \\
\cfrac{\partial L}{\partial v_{ik}} &= \cfrac{\partial L}{\partial \hat{r}} \cdot \cfrac{\partial \hat{r}}{\partial v_{ik}} =  \tanh(\hat{r}(\mathvect{s}) - r(\mathvect{s})) \cdot \left\{\left( \sum_j v_{jk}s_j \right) s_i - v_{ik}\right\}.
\end{align}
Note that in the case of $i=0$ in Eq. 14, $s_i$ is set to +1 since $w_0$ is a bias parameter (see Eq. 1).

Since $\sum_j v_{jk}s_j$ are independent of $i$, the partial derivatives with respect to $v_{ik}$ are calculated efficiently by precomputing $z_k^d$. 
The computational complexity per processing a mini-batch can be reduced to $\mathcal{O}(L_\mathrm{batch}NK)$ by this precomputing \cite{rendle10}.

The parameters ($v_{ik}$ and $w_i$) are updated using backpropagation method, which updates the parameters along with the gradients.
More specifically, the model training unit updates based on Adam optimizer algorithm \cite{kingma2017} shown in Algorithm \ref{algo_adam}.
The Adam optimizer adjusts its learning rate based on the estimations of first-order moment and second-order moment.
The hyper-parameters for Adam optimizer for the RT-BBO implementations are $\beta_1=0.9$, $\beta_2=0.999$, and $\epsilon=10^{-8}$.
The learning rate $\alpha$ for the Adam optimizer is set depending on the kind of parameters and experiment configurations.

\begin{algorithm}[H]
\scriptsize
\tcc{The parameters for quadratic terms $v_{ik}$ are initialized with uniformly random number between -0.001 and 0.001,}
\tcc{and the parameters for linear terms and bias $w_{i}$ are initialized to zero at the first sampling cycle.}
\tcc{The parameters are carried over from the previous sampling cycle otherwise.}
\If{ this sampling cycle is the first sampling cycle}{
\For{$\mathbf{each}\ v_{ik}$}{
    $v_{ik} \leftarrow \mathcal{U}(-0.001, 0.001)$\;
}
\For{$\mathbf{each}\ w_{i}$}{
    $w_{i} \leftarrow 0$\;
}
}

\tcc{Pre-training weight decay. $c_\mathrm{decay}$ is the coefficient for weight decay. }
\For{$\mathbf{each}\ v_{ik}$}{
    $v_{ik} \leftarrow c_\mathrm{decay} v_{ik}$\;
}

\tcc{Parameter Updater\\
The superscripts in the upper right are not exponents, but data indices in the batch.}
\For{$t \leftarrow 1\ \mathbf{to}\ N_\mathrm{train}$}{
    \For{$\mathbf{each\ datum}\ (\mathvect{s}^b, r^b)\ \mathbf{in\ batch}\ (b \leftarrow 1\ \mathbf{to}\ L_\mathrm{batch})$}{
        $\hat{r}^b \leftarrow \sum_i^N \sum_j^N \langle \mathvect{v}_i, \mathvect{v}_j \rangle s_i^b s_j^b + \sum_i^N w_i s_i^b + w_0$\;
        $\cfrac{\partial L^b}{\partial \hat{r}} \leftarrow  - \tanh(r^b - \hat{r}^b)$\;
        \For{$k\ \leftarrow\ 1\ \mathbf{to}\ K$}{
            $z_k^b \leftarrow \sum_j^N v_{jk} s_j^b$\;
        }
    }
    \For{$\mathbf{each\ parameter}\ v_{ik}$}{
        $\cfrac{\partial L}{v_{ik}} \leftarrow \cfrac{1}{L_\mathrm{batch}} \sum_{b=1}^{L_\mathrm{batch}} \cfrac{\partial L^b}{\partial \hat{r}} (z_k^b s_i^b - v_{ik})$\;
        Update parameter $v_{ik}$ using Adam optimizer with gradient $\cfrac{\partial L}{v_{ik}}$\;
    }
    \For{$\mathbf{each\ parameter}\ w_{i}$}{
        $\cfrac{\partial L}{w_{i}}  \leftarrow \cfrac{1}{L_\mathrm{batch}} \sum_{b=1}^{L_\mathrm{batch}} \cfrac{\partial L^b}{\partial \hat{r}} s_i^b$\;
        Update parameter $w_i$ using Adam optimizer with gradient $\cfrac{\partial L}{w_{i}}$\;
    }
}
\caption{Algorithm for the model training unit}
\label{algo_mtu}
\end{algorithm}

\begin{algorithm}[H]
\scriptsize
\SetAlgoLined
\KwIn{Parameter $\theta$, moment $m$, velocity $v$, gradient $g$, coefficients $\beta_1$, $\beta_2$, $\alpha$, $\epsilon$}
\KwOut{Updated parameter $\theta$, moment $m$, velocity $v$}
\tcc{The moments and velocities are maintained for each parameter\\ and initialized to zero at the beginning of the first sampling cycle.}
$m \leftarrow \beta_1 \cdot m + (1 - \beta_1) \cdot g$\;
$v \leftarrow \beta_2 \cdot v + (1 - \beta_2) \cdot g^2$\;
$\hat{m} \leftarrow m / (1 - \beta_1^t)$\;
$\hat{v} \leftarrow v / (1 - \beta_2^t)$\;
$\Delta \theta \leftarrow \alpha \cdot \hat{m} / (\sqrt{\hat{v}} + \epsilon)$\;
$\theta \leftarrow \theta - \Delta \theta$;\tcc{In the case of conventional weight decay, $\theta \leftarrow c_\mathrm{decay}\theta - \Delta \theta$}
\caption{Adam optimizer used in the parameter updater}
\label{algo_adam}
\end{algorithm}

\subsection*{Implementation}

\begin{figure}[tb]
\centering
\includegraphics[width=11.2 cm]{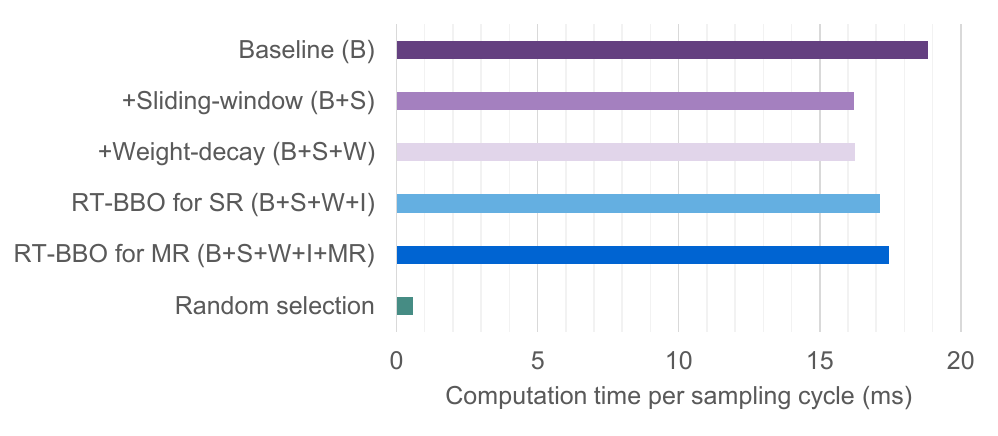}
\caption{
Comparison of computation time per sampling cycle. 
The baseline needs longer time than the others because the data used for training are randomly selected from the large dataset, whereas the FMSBs use the limited data in the smaller sliding-window dataset in the recorded order. The calculation of incentive takes additional time of 1 ms. As for the multi-reward modeling, the increase in computation time is relatively small because the parallel processing on the GPU and the reduction of data transfers work effectively.
Note that each computation time includes the computation time for the wireless simulation (corresponds to the time of random selection).
}
\label{Fig_time}
\end{figure}

The FMSBs, including the RT-BBO used in our experiments, are implemented on a server equipped with both an FPGA board and a GPU board.
The GPU serves as a machine learning accelerator for the model training unit, while the FPGA serves as an embedded Ising machine.
Higher sampling rates increase the volume of data used to approximate the surrogate model that reflects the current environment, which is expected to enhance optimization performance.
Fig.~\ref{Fig_time} shows the computation time per sampling cycle in the beamforming optimization problem (see Fig. 4).
The RT-BBO for SR achieves 17.1 ms per sampling cycle (58 sampling cycles per second), and the RT-BBO for MR achieves 17.5 ms per sampling cycle (57 sampling cycles per second) for the 19 sub-reward cases.
Note that the original FMQA~\cite{kitai20} is expected to require more time than the baseline, as it updates parameters using the entire dataset without employing mini-batch techniques.

The computation flow in a single sampling cycle is as follows.
First, the CPU receives the latest sampling datum from the environment and stores it in the sliding-window dataset.
The CPU also computes the incentive and transfers both the dataset and the incentive to the GPU.
The GPU then updates the surrogate model through the model training unit, which applies the pre-training weight decay followed by parameter updates using the Adam optimizer.
In the case of multiple instant rewards, all sub-surrogate models are trained in parallel and integrated into a single surrogate model.
Next, the acquisition function is computed on the GPU as the sum of the surrogate model output, the incentive, and penalty terms for one-hot encoding.
This acquisition function is transferred via main memory to the FPGA in the form of an Ising model.
The embedded Ising machine on the FPGA computes an approximated solution to the Ising model.
Finally, the CPU simulates the environment using the action corresponding to the obtained solution.

The hardware includes an NVIDIA GeForce RTX 4060 Ti GPU with 4352 CUDA cores (driver version: 550.90.07), an Intel FPGA Programmable Acceleration Card D5005 with a 2800K-LE Stratix 10 FPGA, and two Intel(R) Xeon(R) E5-2667 v4 CPUs. The model training unit is implemented in CUDA (version 12.4) and runs on the GPU. The FPGA implementation of the SB-based Ising machine is the same as our previous work \cite{Hidaka23}.
The conversion between discrete and spin variables, the calculation of constraint terms, and of the incentive computations are implemented in Python, while the remaining components are written in C++.

In the demonstration, the sliding-window size is set to 50, and the coefficient for the pre-training weight decay is set to 0.999.
The range to adjust the coefficient for the incentive is set to 100--200 (see Sec. 2.1.4).
To adjust the dynamic range of the instant rewards, we scale the rewards by a coefficient before adding them to the dataset: by a factor of $1,000$ for the RT-BBO for MR, and by $100,000/19$ for the RT-BBO for SR.

\subsection*{Wireless Environment Simulator}
The simulation of wireless environment is based on the equations and descriptions in Ref.~\cite{DRL_for_DDBC}. The self-correlation coefficient $\rho$ representing the strength of Rayleigh fading is set to 0.90, which is the level corresponding to sampling interval of 10 ms, movement speed of 5 km/h, and communication frequency of 2 GHz in an ideal environment \cite{dong04}. In the cellular network, the cell radius is 200 m, the number of antenna elements per base station is 3, and the number of multipaths (for Rayleigh fading) is 4. The attenuation based on communication distance is set to $120.9 + 37.6 \log_{10} d$ dB, where $d$ is the distance between the user and the base station in km. The environment simulator was written in C++ and runs on the CPU.

\section*{Data availability}
The authors declare that all relevant data are included in the manuscript and the Supplementary information. Additional data are available from the corresponding author upon reasonable request.

\section*{Code availability}
The extensions made in this work, relative to the baseline (FMQA~\cite{kitai20}), and the hyperparameter values are detailed in the main text through pseudocodes and mathematical equations.

\section*{Acknowledgements}
The authors would like to thank Daisuke Uchida, Kentaro Taniguchi, and Duckgyu Shin for their advice.

\section*{Competing interests}
T.K., Y.H., M.Y., and K.T. are included as inventors on two patent applications related to this work filed by the Toshiba Corporation in Japan (P2024-018133 and P2025-007789).
The authors declare that they have no other competing interests.

\section*{Author contributions}
All authors collaboratively discussed and determined the direction of this work.
T.K. and K.T. conceived the idea of extending FMQA to real-time systems.
K.T., Y.H., and M.Y. managed this project and advised on the SB-based Ising machine.
T.K. devised the algorithm of RT-BBO, implemented the RT-BBO and the wireless simulator, and evaluated them.
T.K. and K.T. wrote this manuscript. 

\section*{Additional information}

\subsection*{Supplementary information}

The online version contains supplementary materials.
\begin{itemize}
\item Supplementary information 1: A document that explains the Supplementary information 2.
\item Supplementary information 2: A video demonstrating the beamforming states and achieved throughputs of the proposed RT-BBO and greedy methods for dynamic optimization of a wireless cellular network.
\end{itemize}
\textbf{Note that the preprint (arXiv) version does not contain the Supplementary information 2.}

\printbibliography

\newpage

\section*{Supplementary Information 1}
Supplementary Information 2 presents a visual demonstration of dynamic beamforming optimization using the proposed RT-BBO for MR in a wireless cellular network. This video illustrates the process of 36,000 sampling iterations, corresponding to a single trial described in the Demonstration section. The results of the RT-BBO for MR are compared with the greedy method described in the Main text.

The top half of the video shows the users' locations and the states of the beams evaluated during sampling, while the bottom half displays the communication throughput. Each plot point represents the communication throughput achieved at a sampling step, with the solid line indicating the moving average.

The greedy method selects the beamforming pattern that maximizes the received power for the corresponding user based on environmental observations (using internal information from the black box function), without considering interference from other base stations. In contrast, the RT-BBO accounts for the effects of interference and achieves higher throughput than the greedy method most of the time.

\begin{figure}[hb]
\centering
\fbox{
\includegraphics[width=15.0 cm]{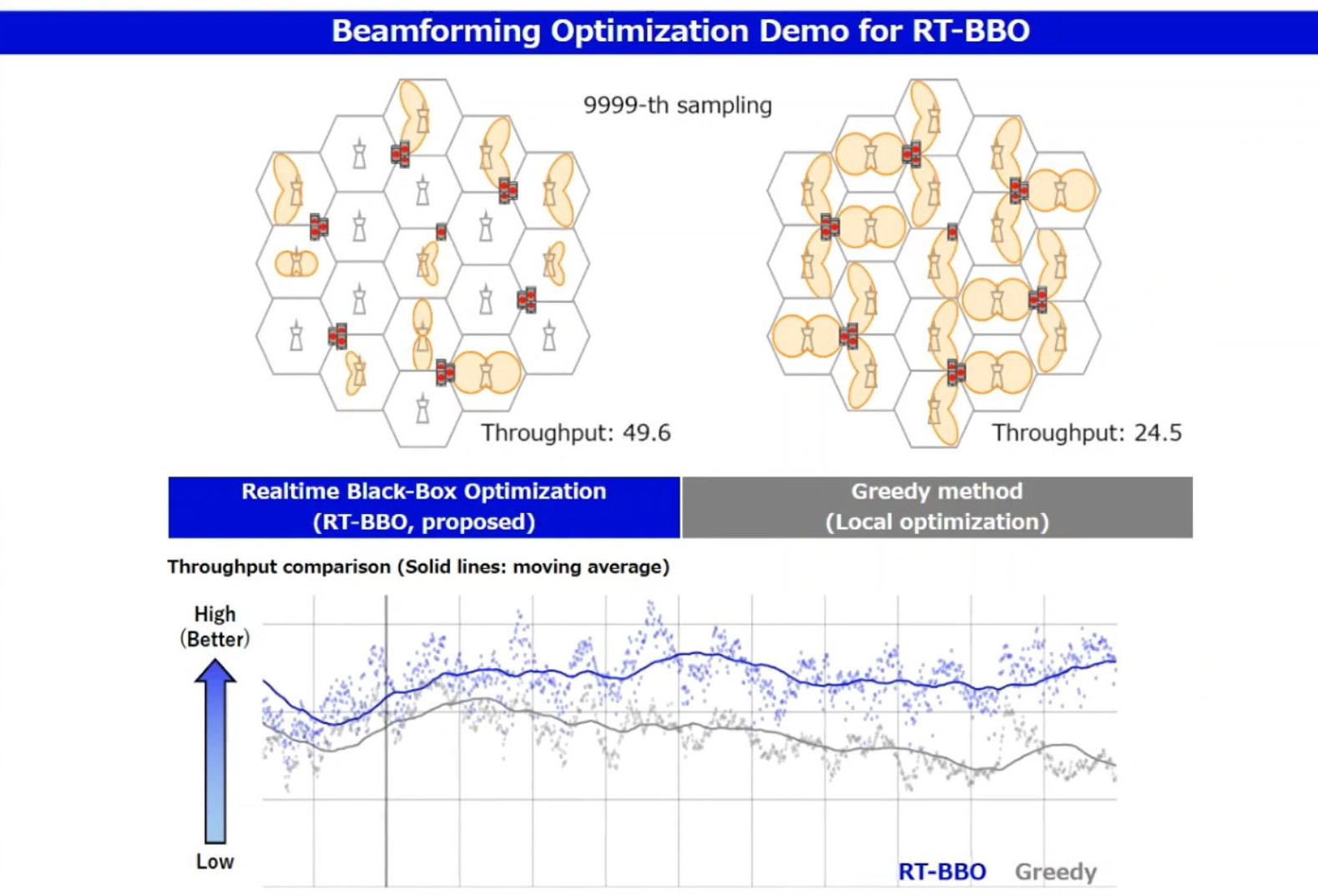}
}
\caption{A snapshot from the Supplementary information 2}
\label{Fig_video_snapshot}
\end{figure}

\end{document}